\documentclass[letterpaper, 10pt, conference]{ieeeconf} % Comment this line out if you need a4paper

\IEEEoverridecommandlockouts % This command is only needed if you want to use the \thanks command

\usepackage{cite}
\usepackage{amsmath,amssymb,amsfonts}
\usepackage{algorithmic}
\usepackage{graphicx}
\usepackage{textcomp}
\usepackage{xcolor}
\def\BibTeX{{\rm B\kern-.05em{\sc i\kern-.025em b}\kern-.08em
    T\kern-.1667em\lower.7ex\hbox{E}\kern-.125emX}}
\usepackage{booktabs}
\usepackage{multirow}
\usepackage[super]{nth}
\usepackage{tikz}
\newcommand*\circled[1]{\tikz[baseline=(char.base)]{
		\node[shape=circle,draw,inner sep=0.2pt] (char) {#1};}}

\usepackage{soul}

\usepackage{setspace}
\usepackage[urlcolor=blue]{hyperref} 
\usepackage{cleveref}

\usepackage{fancyhdr}
\fancypagestyle{firstpage}{
  \fancyhf{}
  \chead{\small To appear at the 18th International Conference on Control, Automation, Robotics and Vision (ICARCV), December 2024, Dubai, UAE.}
  \cfoot{\thepage}
}

\title{\LARGE \bf A Methodology to Study the Impact of Spiking Neural Network Parameters considering Event-Based Automotive Data
\vspace{-0.2cm}}

\author{Iqra Bano*, Rachmad Vidya Wicaksana Putra*, Alberto Marchisio, Muhammad Shafique% <-this % stops a space
\thanks{Iqra Bano, Rachmad Vidya Wicaksana Putra, and Alberto Marchisio are with eBrain Lab, Division of Engineering, New York University (NYU) Abu Dhabi, United Arab Emirates;
{e-mail: \{ib2419, rachmad.putra, alberto.marchisio\}@nyu.edu}
}%
\thanks{Muhammad Shafique is the Director of eBrain Lab, Division of Engineering, New York University (NYU) Abu Dhabi, United Arab Emirates;
{e-mail: muhammad.shafique@nyu.edu}
}
\thanks{*These authors have equal contributions.}% <-this % stops a space
\vspace{-0.4cm}
}

\begin{document}

\maketitle
\pagestyle{plain}
\thispagestyle{firstpage}

%%%%%%%%%%%%%%%%%%%%%%%%%%%%%%%%%%%%%%%%%%%%%%%%%%
%%%%%%%%%%%%%%%%%%%%%%%%%%%%%%%%%%%%%%%%%%%%%%%%%%
\begin{spacing}{1}
\begin{abstract}
Autonomous Driving (AD) systems are considered as the future of human mobility and transportation. 
Solving computer vision tasks such as image classification and object detection/segmentation, with high accuracy and low power/energy consumption, is highly needed to realize AD systems in real life.
These requirements can potentially be satisfied by Spiking Neural Networks (SNNs). 
However, the state-of-the-art works in SNN-based AD systems still focus on proposing network models that can achieve high accuracy, and they have not systematically studied the roles of SNN parameters when used for learning event-based automotive data. 
Therefore, we still lack understanding of how to effectively develop SNN models for AD systems. 
Toward this, we propose a novel methodology to systematically study and analyze the impact of SNN parameters considering event-based automotive data, then leverage this analysis for enhancing SNN developments.
To do this, we first explore different settings of SNN parameters that directly affect the learning mechanism (i.e., batch size, learning rate, neuron threshold potential, and weight decay), then analyze the accuracy results. 
Afterward, we propose techniques that jointly improve SNN accuracy and reduce training time. 
Experimental results show that our methodology can improve the SNN models for AD systems than the state-of-the-art, as it achieves higher accuracy (i.e., 86\%) for the NCARS dataset, and it can also achieve iso-accuracy (i.e., $\sim$85\% with standard deviation less than 0.5\%) while speeding up the training time by 1.9x. 
In this manner, our research work provides a set of guidelines for SNN parameter enhancements, thereby enabling the practical developments of SNN-based AD systems. 
\end{abstract}
%
% \begin{IEEEkeywords}
% Spiking neural networks, neuromorphic computing, autonomous driving, event-based automotive data, event-based cameras, NCARS dataset.
% \end{IEEEkeywords}

%%%%%%%%%%%%%%%%%%%%%%%%%%%%%%%%%%%%%%%%%%%%%%%%%%
%%%%%%%%%%%%%%%%%%%%%%%%%%%%%%%%%%%%%%%%%%%%%%%%%%
\section{Introduction}
\label{Sec_Intro}

In recent years, the interest in the autonomous driving (AD) systems has increased significantly as these systems potentially improve the efficiency of human mobility and transportation~\cite{Ref_Viale_CarSNN_IJCNN21}\cite{Ref_Cordone_ObjDetSNN_IJCNN22}.
The development of AD systems in real life necessitates the capabilities of solving computer vision tasks such as image classification and object detection/segmentation from images/videos. 
Currently, the state-of-the-art accuracy for computer vision tasks is achieved through advanced neural network algorithms, such as Deep Neural Networks (DNNs) and Spiking Neural Networks (SNNs). 
Therefore, the employment of neural network algorithms for AD systems is prevalent~\cite{Ref_Shafique_EdgeAI_ICCAD21}.
Besides the above-discussed functionalities, AD systems also require low energy consumption to preserve their battery lifespan, and real-time output to provide fast decision.
These requirements can potentially be fulfilled by SNNs, since SNNs offer low power/energy computation due to sparse event-based operations~\cite{Ref_Putra_Mantis_ICARA23, Ref_Tavanaei_DLSNN_Neunet18, Ref_Putra_FSpiNN_TCAD20}, high accuracy due to effective learning mechanism~\cite{Ref_Viale_CarSNN_IJCNN21}\cite{Ref_Cordone_ObjDetSNN_IJCNN22}\cite{Ref_Viale_LaneSNN_IROS22}\cite{Ref_Putra_SpikeDyn_DAC21}, and low latency due to efficient neural/spike coding and operational timesteps~\cite{Ref_Guo_NeuralCoding_FNINS21, Ref_Park_T2FSNN_DAC20, Ref_Chowdhury_LowLatencySNNs_ECCV22, Ref_Putra_TopSpark_IROS23}.
Furthermore, to maximize the benefits of sparse operations in SNNs, the workload should be presented in the form of event-based data, because this data format is suitable with the data representation in SNNs (i.e., spikes).
Therefore, to ensure the practicality of the SNN models for AD systems, the SNN developments should also consider event-based automotive data, such as the NCARS dataset~\cite{Ref_Sironi_HATS_CVPR18}. 
Motivated by the potentials of SNN-based AD systems, \textbf{our targeted research problems} are the following.
\begin{itemize}
    \item \textit{How can we investigate, analyze, and understand the impact of different SNN parameters on final accuracy?}
    \item \textit{How can we effectively leverage this analysis and the obtained knowledge for developing SNN models for event-based automotive data?}
\end{itemize} 
The efficient solution to this problem will provide guidelines and insights for developing effective SNN models that can achieve high accuracy with fast training time for AD systems.
To address these problems, we make two important contributions in this paper: (1) \textit{detailed investigation and analysis on how different SNN parameters and settings impact the accuracy}, and (2) \textit{a novel methodology to develop SNN models considering the key observations from the above analysis}; which will be discussed further in Section~\ref{Sec_Intro_Novelty}.

%%%%%%%%%%%%%%
\subsection{State-of-the-Art and Their Limitations}
\label{Sec_Intro_SOTA}

\begin{figure}[t]
\centering
\includegraphics[width=0.92\linewidth]{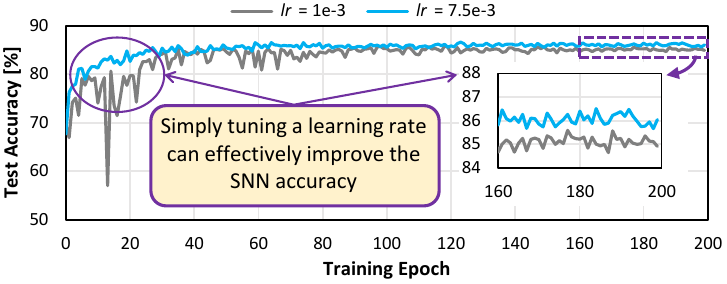}
\vspace{-0.3cm}
\caption{A simple tuning to the learning rate ($lr$) of the state-of-the-art CarSNN work~\cite{Ref_Viale_CarSNN_IJCNN21} from its default setting can effectively improve the SNN accuracy for event-based automotive data (i.e., NCARS dataset).
The accuracy improvements happen over the training phase, leading to higher accuracy with smaller training time.} 
\label{Fig_Observe_Default}
\vspace{-0.6cm}
\end{figure}

Currently, the state-of-the-art works in SNN-based AD systems~\cite{Ref_Viale_CarSNN_IJCNN21}\cite{Ref_Cordone_ObjDetSNN_IJCNN22} still focus on proposing network models to achieve high accuracy on event-based automotive data.
However, \textit{they have not investigated the impact and roles of different SNN parameters on the learning quality, and hence lack understanding on how to effectively develop SNN models for AD systems.} 
Consequently, the existing techniques for developing SNNs for AD systems may not be efficient, and their accuracy may be sub-optimal.    
To demonstrate these limitations, we perform an experimental case study.
Here, we implement CarSNN~\cite{Ref_Viale_CarSNN_IJCNN21} and perform experiments with its default parameter setting.
Note, the details for the experimental setup and parameter settings will be discussed in Sections~\ref{Sec_EvalMethod} and~\ref{Sec_Results}.
The experimental results with key observations are presented in Fig.~\ref{Fig_Observe_Default}.

These results show that with a simple tuning of the learning rate, the SNN accuracy for the NCARS dataset is improved, and high accuracy can be obtained with fewer training epochs compared to the state-of-the-art. 
These limitations expose the need to understand the impact of SNN parameters when learning event-based automotive data so that the knowledge can be used for practical SNN developments for AD systems. 

\textit{\textbf{Required:}
An understanding of the impact of changing different SNN parameters on the learning quality, and how to leverage them for developing efficient SNN models for AD systems}. 
However, this requirement exposes several research challenges as summarized in the following.
\begin{itemize}
    \item The investigation should only consider parameters that are important for the learning process (e.g., batch size, learning rate, threshold potential, and weight decay), so that the investigation does not need to consider all SNN parameters. 
    \item Different SNN parameters may need different types of investigation and enhancement strategies due to their diverse roles and functionalities.  
\end{itemize}

%%%%%%%%%%%%%%
\vspace{-0.2cm}
\subsection{Our Novel Contributions}
\label{Sec_Intro_Novelty}
\vspace{-0.1cm}

\begin{figure}[t]
	\centering
	\includegraphics[width=0.9\linewidth]{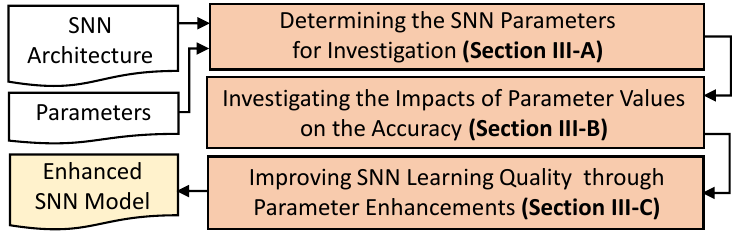}
	\vspace{-0.3cm}
	\caption{An overview of our novel contributions, shown in orange boxes.} 
	\label{Fig_NovelContrib}
	\vspace{-0.6cm}
\end{figure}

To address the above-discussed problems and challenges, we propose \textit{a novel methodology to systematically analyze the impact of SNN parameters on the learning quality, and then leverage this analysis for devising effective SNN architectures for AD systems}. 
Its key steps are presented in the following (see overview in Fig.~\ref{Fig_NovelContrib}).
\begin{itemize}
    \item \textbf{Identifying the SNN parameters for investigation (Section~\ref{Sec_Method_SelectedParams}):} 
    This step selects parameters that are important for the learning process. 
    Here, we select batch size, learning rate, neurons' threshold potential, and weight decay. 
    \item \textbf{Investigating the impact of parameter values on the accuracy (Section~\ref{Sec_Method_Xplore}):}
    This step explores different values of the selected SNN parameters, and analyzes the accuracy. 
    \item \textbf{Improving SNN learning quality based on our parameter enhancements (Section~\ref{Sec_Method_Improve}):} 
    This step leverages the knowledge from previous analysis by tuning the parameter settings to improve the SNN accuracy and/or reduce the training time, thus leading to better SNNs for AD systems.
\end{itemize}  
\textbf{Key results:}
Our proposed methodology is evaluated through a Python-based implementation that runs on the Nvidia RTX 6000 Ada GPU machine with the NCARS dataset as workload. 
The experimental results show that our methodology improves the SNN models from the state-of-the-art, i.e., by achieving higher accuracy (i.e., 86\%), and achieving iso-accuracy\footnote{Iso-accuracy means the same or similar accuracy values between two instances. In this work, we consider iso-accuracy if two accuracy scores have the same integer value and/or have less than 1\% accuracy difference.} (i.e., $\sim$85\% with standard deviation $<$ 0.5\%) while speeding up the training phase by 1.9x. 

%%%%%%%%%%%%%%%%%%%%%%%%%%%%%%%%%%%%%%%%%%%%%%%%%%
%%%%%%%%%%%%%%%%%%%%%%%%%%%%%%%%%%%%%%%%%%%%%%%%%%
% \vspace{-0.1cm}
\section{Background}
\label{Sec_Back}

%%%%%%%%%%%%%%
\subsection{Spiking Neural Networks (SNNs)}
\label{Sec_Back_SNNs}

\begin{figure}[hbtp]
\vspace{-0.2cm}
\centering
\includegraphics[width=\linewidth]{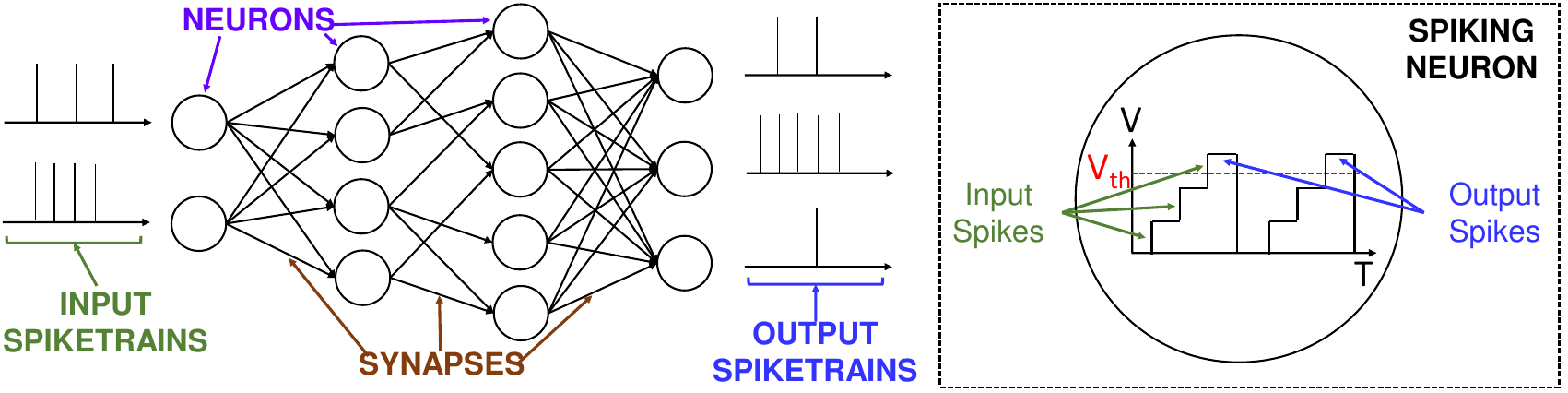}
\vspace{-0.6cm}
\caption{Overview of the functionality of a Spiking Neural Network.} 
\label{Fig_SNN}
\vspace{-0.4cm}
\end{figure}

Recently, SNNs have demonstrated great success and potential to be deployed in autonomous systems due to their low energy consumption and high performance. Inspired by nature and biology, the artificially-designed SNNs mimic the event-based decision-making process of the biological brain. As shown in Fig.~\ref{Fig_SNN}, they receive input in the form of a spike train and process the information through spiking neurons.
SNNs typically employ a specific neuron model, such as the Leaky Integrate-and-Fire (LIF), whose neuronal dynamics can be stated as the following.
\begin{equation}
    \begin{split}
    \tau \frac{dV_{m}(t)}{dt} &  = - (V_{m}(t)-V_{r}) + I(t) 
    \end{split}
    \label{Eq_LIF}
\end{equation}
\begin{equation}
    \begin{split}
    \text{if} \;\;\; V_{m} & \geq V_{th} \;\;\; \text{then} \;\;\; V_{m} \leftarrow V_{r}
    \end{split}
    \label{Eq_LIF_fire}
\end{equation}
Here, $V_{m}$ refers to the membrane potential of a neuron at time-$t$, $V_{r}$ refers to the reset potential of a neuron, and $\tau$ refers to the time constant of membrane potential decay. 
$I$ refers to the inputs. 
Meanwhile, $V_{th}$ refers to the neurons' threshold potential (or simply threshold potential).

Compared to conventional DNNs, SNNs have advantages in terms of low power/energy consumption, low computation latency, as well as a straightforward interface with event-based cameras. 
Based on these benefits, the deployment of SNNs for AD systems is highly desired. 

%%%%%%%%%%%%%%
\subsection{SNN Learning Mechanism}
\label{Sec_Back_Learning}

\textbf{Overview:}
Training SNNs is non-trivial due to the intrinsic non-differentiable nature of the spiking loss function~\cite{Ruckauer_2019arxiv_NonDifferentiableLossFunctionSNN}. 
To overcome this issue, there are two possible solutions proposed in the literature for training SNNs with supervised settings, namely \textit{DNN-to-SNN conversion}~\cite{Massa_2020IJCNN_EfficientSNNGestures} and \textit{direct SNN training under surrogate gradient}~\cite{Ref_Neftci_SurrogateSNNs_IEEEMSP19}. 
Since the first approach requires training non-spiking DNNs on the same data, it limits its applicability to static data and cannot be directly used on event-based data. 
Meanwhile, the second approach trains the network directly in the spiking domain through a surrogate gradient-based learning rule to address the non-differentiable issue in SNNs.
In this category, the Spatio-Temporal Back-Propagation~\cite{Ref_Wu_STBP_FNINS18} is one of the prominent works. 

\textbf{Spatio-Temporal Back-Propagation (STBP)~\cite{Ref_Wu_STBP_FNINS18}:} 
STBP learning rule employs the information from the spatio-temporal domain (STD). 
The STBP rule uses both the spatial domain (SD) and temporal domain (TD). It updates the membrane potential according to Eq.~\ref{eq_Vm_STBP}, where $I(t)$ represents the spatial accumulation and $V_m(t_{i-1})$ represents the leaky temporal memory. In this way, the membrane potential decays until the neuron receives another input spike, which triggers another update round.

\begin{equation}
V_m(t)=V_m(t_{i-1})e^{\frac{t_{i-1}-t}{\tau}}+RI(t)
\label{eq_Vm_STBP}
\end{equation}

This allows to iteratively represent the chain rule updating the pre-synaptic input $x_i$, the membrane potential $V_{mi}$ and the neuronal output $o_i$ of the $i^{th}$ neuron according to Eq.~\ref{eq_system_STBP_model}. 
Here, the index $t$ denotes the current timestep, $n$ and $l(n)$ represent the $n^{th}$ layer and its number of neurons, respectively, $w_{ij}$ is the weight between the pre-synaptic neuron $j$ and the post-synaptic neuron $i$, and $b_i$ is the bias of the neuron $i$. 
Additionally, $f$ and $g$ are specific functions described in Eq.~\ref{eq_system_STBP_model_f} and Eq.~\ref{eq_system_STBP_model_g}, respectively.

\begin{equation}
\begin{cases}
x_{i}^{t+1, n} &=\sum_{j=1}^{l(n-1)} w_{i j}^{n} o_{j}^{t+1, n-1} \\
V_{mi}^{t+1, n} &=V_{mi}^{t, n} f\left(o_{i}^{t, n}\right)+x_{i}^{t+1, n}+b_{i}^{n} \\
o_{i}^{t+1, n} &=g\left(V_{mi}^{t+1, n}\right),
\end{cases}
\label{eq_system_STBP_model}
\end{equation}

\begin{equation}
f\left(o_{i}^{t, n}\right) \approx\left\{\begin{array}{ll}
\tau, & o_{i}^{t, n}=0 \\
0, & o_{i}^{t, n}=1
\end{array}\right.\ for\ small\ \tau
\label{eq_system_STBP_model_f}
\end{equation}

\begin{equation}
g(x)=\left\{\begin{array}{ll}
1, & x \geq V_{t h} \\
0, & x<V_{t h}
\end{array}\right.
\label{eq_system_STBP_model_g}    
\end{equation}

The loss function for the STBP is defined in Eq.~\ref{eq_loss_STBP}, where $y_s$ is the label and $o_s$ is the output of the $s_{th}$ training
sample.
\begin{equation}
L=\frac{1}{2 S} \sum_{s=1}^{S}\left\|y_{s}-\frac{1}{T} \sum_{t=1}^{T} o_{s}^{t, N}\right\|_{2}^{2}
\label{eq_loss_STBP}    
\end{equation}
Afterward, the gradients are computed as the partial derivatives of the loss with respect to the outputs, according to Eq.~\ref{eq_system_STBP_gradients}. 
The non-differentiable function $g$ is approximated through the surrogate model from the studies in~\cite{Ref_Neftci_SurrogateSNNs_IEEEMSP19}.

\begin{equation}
\delta_{i}^{t, n}=\frac{\partial L}{\partial o_{i}^{t, n}}    \label{eq_system_STBP_gradients}
\end{equation}

%%%%%%%%%%%%%%
\subsection{SNN Architecture}
\label{Sec_Back_NetArch}

The SNN architecture considered in this work is based on the studies in~\cite{Ref_Viale_CarSNN_IJCNN21}, and supports the STBP-based learning mechanism. 
The architectural details are provided in the Table~\ref{Tab_SNNarch}. It is composed of two convolutional layers with 32 output channels, $3 \times 3$ kernel size, padding of 1, and stride of 1, interleaved by average pooling layers, and followed by two fully-connected layers (dense layers). 

\begin{table}[hbtp]
\caption{SNN architecture considered in this work.}
\label{Tab_SNNarch}
\vspace{-0.2cm}
\scriptsize
\centering
\begin{tabular}{|c|c|c|c|c|c|}
\hline
\textbf{\begin{tabular}[c]{@{}c@{}}Layer\\ Type\end{tabular}} & \textbf{\begin{tabular}[c]{@{}c@{}}Input \\ Channel\end{tabular}} & \textbf{\begin{tabular}[c]{@{}c@{}}Output\\ Channel\end{tabular}} & \textbf{\begin{tabular}[c]{@{}c@{}}Kernel\\ Size\end{tabular}} & \textbf{Padding} & \textbf{Stride} \\ \hline \hline
Avg. Pooling & 2 & 2 & 4 & - & - \\ \hline
Convolution & 2 & 32 & 3 & 1 & 1 \\ \hline
Avg. Pooling & 32 & 32 & 2 & - & - \\ \hline
Convolution & 32 & 32 & 3 & 1 & 1 \\ \hline
Avg. Pooling & 32 & 32 & 2 & - & - \\ \hline
Dense & 1152 & 512 & - & - & - \\ \hline
Dense & 512 & 2 & - & - & - \\ \hline
\end{tabular}
\vspace{-0.4cm}
\end{table}

%%%%%%%%%%%%%%
\subsection{Event-based Automotive Data}
\label{Sec_Back_EventAutoData}

\begin{figure}[hbtp]
\vspace{-0.3cm}
\centering
\includegraphics[width=0.82\linewidth]{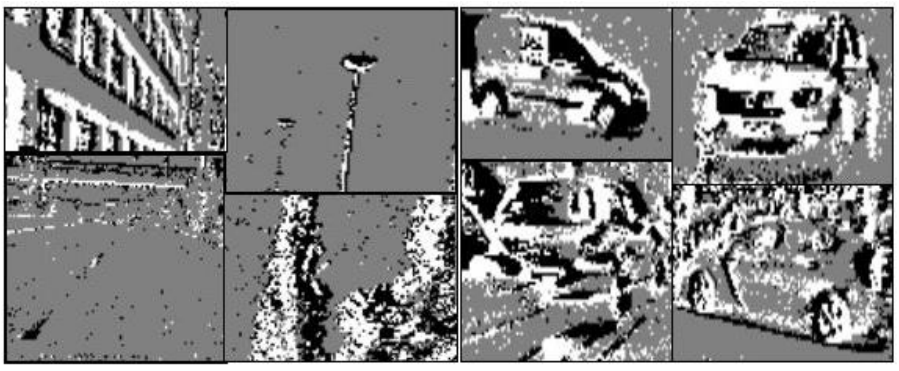}
\vspace{-0.3cm}
\caption{Example of images from the NCARS dataset~\cite{Ref_Viale_CarSNN_IJCNN21}\cite{Ref_Sironi_HATS_CVPR18}.} 
\label{Fig_NCARS_dataset}
\vspace{-0.3cm}
\end{figure}

\textbf{Prophesee NCARS Dataset~\cite{Ref_Sironi_HATS_CVPR18}:}
This dataset has around 24K samples, and each having a 100ms length.
The samples represent either ``background'' or ``car'' and are obtained through recordings using the Asynchronous Time-based Image Sensor (ATIS) camera~\cite{Ref_Sironi_HATS_CVPR18}.
The input samples are a sequence of events that encode the following information:
\begin{itemize}
    \item the spatial coordinates $x,y$ of the pixel;
    \item the timestamp $t$ of when the event occurred;
    \item the polarity $p$ of the brightness variation, which can be either positive or negative.
\end{itemize}
These samples are split into 7940 training and 4396 testing samples for car, as well as 7482 training and 4211 testing samples for background. A few examples are shown in Fig.~\ref{Fig_NCARS_dataset}.
These samples have variable sizes as they are typically cropped from the original recordings (i.e., resolution of $304 \times 240$ pixels)~\cite{Ref_Cordone_ObjDetSNN_IJCNN22}. 
In this work, we consider $100 \times 100$ pixel resolution as it is sufficient to provide important information~\cite{Ref_Viale_CarSNN_IJCNN21}.

%%%%%%%%%%%%%%%%%%%%%%%%%%%%%%%%%%%%%%%%%%%%%%%%%%
%%%%%%%%%%%%%%%%%%%%%%%%%%%%%%%%%%%%%%%%%%%%%%%%%%
\vspace{-0.2cm}
\section{The Proposed Methodology}
\label{Sec_Method}

\begin{figure*}[t]
\centering
\includegraphics[width=0.9\linewidth]{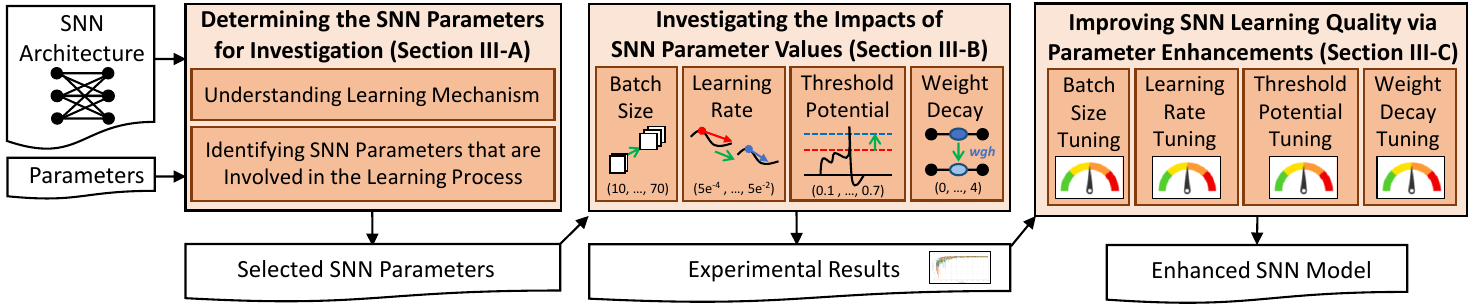}
\vspace{-0.4cm}
\caption{An overview of our proposed methodology and the novel contributions are shown in orange-colored boxes.} 
\label{Fig_Method}
\vspace{-0.5cm}
\end{figure*}

Our proposed methodology systematically investigates the impact of SNN parameters on the learning quality, and then leverages the knowledge for effective SNN developments for AD systems. 
Its overview is shown in Fig.~\ref{Fig_Method}, and the details of each step will be discussed in the subsequent subsections.

%%%%%%%%%%%%%%
\subsection{Determining the SNN Parameters for Investigation}
\label{Sec_Method_SelectedParams}

\textit{It aims at selecting parameters that are considered important (effective) for the learning process} 
To do this, we first study the learning mechanism employed in SNN processing, then identify SNN parameters that are directly involved and/or effectual in the learning process.

In this work, we employ the STBP learning rule~\cite{Ref_Wu_STBP_FNINS18}, which considers both spatial and temporal information. 
Here, the amount of information employed for updating the weights during the learning process mainly depends on the batch size ($B$). 
To calculate its loss function, the STBP rule leverages the spiking activity which mainly depends on the threshold potential ($V_{th}$).  
The reason is that, the threshold potential defines the range of value for neuronal dynamics to trigger the spike generation~\cite{Ref_Putra_lpSpikeCon_IJCNN22}. 
If the threshold potential is low, the corresponding neuron can generate a high number of spikes as the membrane potential can reach threshold potential with a small number of input spikes.
On the other hand, if the threshold potential is high, the corresponding neuron may generate a low number of spikes as the membrane potential can only reach threshold potential with a high number of input spikes. 
Furthermore, the STBP rule also employs an Adam optimization technique for minimizing the loss, whose function is dependent on the learning rate ($lr$) and the weight decay ($w_{decay}$)~\cite{Ref_Kingma_Adam_ICLR15}.   
Therefore, in this work, we consider exploring  \textit{\textit{`batch size' $B$}, \textit{`learning rate' $lr$}, \textit{`threshold potential' $V_{th}$}}, and \textit{`weight decay' $w_{decay}$} as the selected SNN parameters for investigation.

%%%%%%%%%%%%%%
\subsection{Exploring the Impact of Different SNN Parameter Values}
\label{Sec_Method_Xplore}

\begin{figure}[t]
\centering
\includegraphics[width=0.97\linewidth]{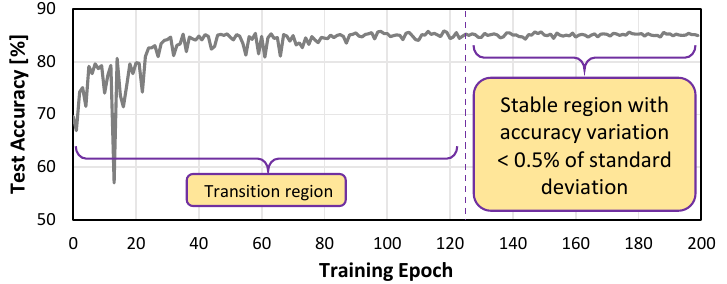}
\vspace{-0.3cm}
\caption{The experimental results of the accuracy under the default setting: $B = 40$, $lr = 1e\text{-}3$, $V_{th} = 0.4$, and $w_{decay} = 0$. 
We observe that the accuracy curve typically consists of two regions, i.e., transition and stable.} 
\label{Fig_Results_Default}
\vspace{-0.6cm}
\end{figure}

\textit{It aims at exploring different settings (values) of the selected SNN parameters, and analyzes their impact on the accuracy}. 
Here, each parameter has its own range of exploration settings due to the nature of their functionalities. 
To decide the range of values for each parameter, we first consider the parameter settings in the state-of-the-art works as the default settings.
For instance, the default setting for the SNN architecture with STBP rule~\cite{Ref_Viale_CarSNN_IJCNN21} considers the following set of values: $B = 40$, $lr = 1e\text{-}3$, $V_{th} = 0.4$, and $w_{decay} = 0$. 
The experimental results for employing the default setting are presented in Fig.~\ref{Fig_Results_Default}.
The results show that, in general, the accuracy curve consists of two regions: (1) \textit{transition region}, where the accuracy is still not stable as the network is still learning new information over training time, and (2) \textit{stable region}, where the accuracy does not change much as the network does not necessarily learn new information.  
This curve indicates how fast the training phase reaches stable accuracy.
We define the stable region as the region that has accuracy with a standard deviation of less than 0.5\% accuracy over the last 10 training epochs, since the typical range of acceptable accuracy variation in the neural network community is within 1\% accuracy~\cite{Ref_Koppula_EDEN_Micro19}\cite{Ref_Putra_SparkXD_DAC21}. 

Afterward, we explore different values for each investigated parameter, which can be smaller or larger than its default setting.
In this manner, we can observe and analyze the correlation between the accuracy and the value changes.
The details of exploration values are defined in the following.
\begin{itemize}
    \item $B \in \{10, 20, 30, 40, 50, 60, 70\}$. 
    \item $lr \in \{5e\text{-}4, 1e\text{-}4, 1e\text{-}3, 5e\text{-}3, 7.5e\text{-}3, 1e\text{-}2, 5e\text{-}2\}$. 
    \item $V_{th} \in \{ 0.1, 0.2, 0.3, 0.4, 0.5, 0.6, 0.7 \}$. 
    \item $w_{decay} \in \{ 0, 0.2, 0.5, 0.75, 1, 2, 4 \}$. 
\end{itemize}

Based on these settings, we perform experiments that tune each investigated parameter with the desired value at one time, while keeping other parameters set with the default values. 
For instance, if we investigate the impact of $B = 10$, then we perform experiments with the following setting: $B = 10$, $lr = 1e\text{-}3$, $V_{th} = 0.4$. 
Note, the experimental results and related discussion for the exploration will be provided in Section~\ref{Sec_Results_BatchSize} until Section~\ref{Sec_Results_Wdecay}.

%%%%%%%%%%%%%%
\subsection{Parameter Enhancements for Improving SNN Learning}
\label{Sec_Method_Improve}

\textit{It aims at improving the SNN learning quality, i.e., increasing the accuracy and/or reducing the training time.}
To do this, we first analyze the experiment results from the previous exploration step to identify the effective value for each parameter. 
The effective parameter value is typically characterized by the one that leads to high accuracy under a relatively short training time. 
Therefore, \textit{our strategy is to perform parameter enhancements by tuning the selected parameters to follow the effective values based on the experimental results.}
Note, the experimental results and related discussion for the SNN parameter enhancements will be provided in Section~\ref{Sec_Results_Enhance} and Section~\ref{Sec_Results_Discuss}.

%%%%%%%%%%%%%%%%%%%%%%%%%%%%%%%%%%%%%%%%%%%%%%%%%%
%%%%%%%%%%%%%%%%%%%%%%%%%%%%%%%%%%%%%%%%%%%%%%%%%%
\section{Evaluation Methodology}
\label{Sec_EvalMethod}

\begin{figure}[h]
\vspace{-0.2cm}
\centering
\includegraphics[width=0.97\linewidth]{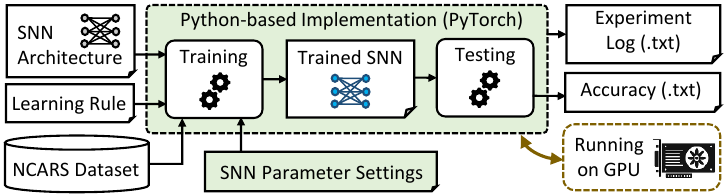}
\vspace{-0.3cm}
\caption{The experimental setup in our evaluation.} 
\label{Fig_EvalMethod}
\vspace{-0.3cm}
\end{figure}

Fig.~\ref{Fig_EvalMethod} illustrates the experimental setup for evaluating our methodology. 
Here, we employ a Python-based implementation that runs on an Nvidia RTX 6000 Ada GPU machine, and the generated outputs are the accuracy (i.e., training and test) and the log of experiments (e.g., number of epoch and loss scores). 
The Python-based implementation is built using the numpy, torch, and torchvision library packages.
We employ the SNN architecture described in Table~\ref{Tab_SNNarch} with the STBP learning rule, the NCARS dataset as the workload, and 200 epochs for the training phase. 
We consider the state-of-the-art work, i.e., CarSNN~\cite{Ref_Viale_CarSNN_IJCNN21} with the following default setting: $B = 40$, $lr = 1e\text{-}3$, $V_{th} = 0.4$, and $w_{decay} = 0$, as the reference comparison partner.

%%%%%%%%%%%%%%%%%%%%%%%%%%%%%%%%%%%%%%%%%%%%%%%%%%
%%%%%%%%%%%%%%%%%%%%%%%%%%%%%%%%%%%%%%%%%%%%%%%%%%
\section{Experimental Results and Discussion}
\label{Sec_Results}

%%%%%%%%%%%%%%
\subsection{Impact of the Batch Size}
\label{Sec_Results_BatchSize}

\begin{figure}[hbtp]
\centering
\includegraphics[width=0.97\linewidth]{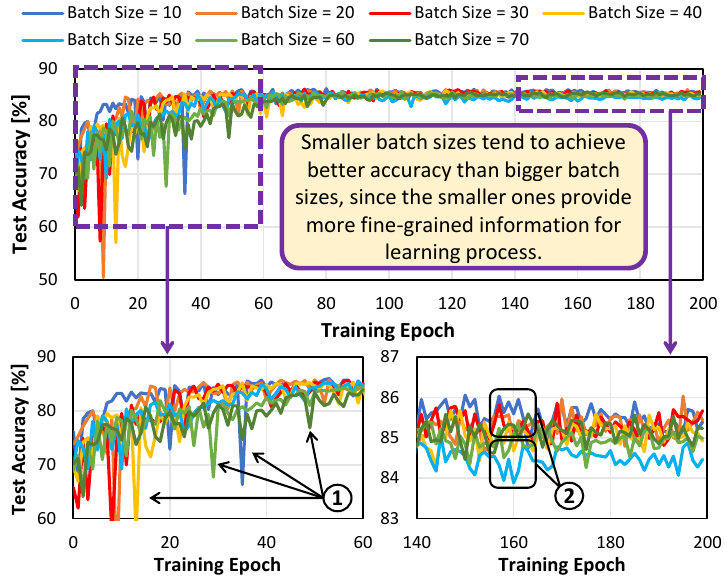}
\vspace{-0.3cm}
\caption{Experimental results for the accuracy across different batch sizes, i.e., $B \in \{10, 20, 30, 40, 50, 60, 70\}$. Here, we observe that smaller batch sizes tend to achieve better accuracy than bigger batch sizes.} 
\label{Fig_Results_BatchSize}
\vspace{-0.3cm}
\end{figure}

The experimental results for the impact of batch size $B$ are presented in Fig.~\ref{Fig_Results_BatchSize}.
The results show that, in general, most of the batch sizes have fluctuating accuracy during the early training phase (i.e., $\leq$ 60 epoch); see label~\circled{1}. 
Such accuracy fluctuations happen because in the early training phase, the network still starts learning new information and needs to jump out of local minima to reach better points in the SNN loss landscape.
Furthermore, we observe that networks with smaller batch sizes tend to face more fluctuations than the ones with bigger batch sizes.
The reason is that, updating the weights under a smaller batch size may lead the network to update its weights towards specific training directions in the early training phase, which may not fully represent unseen features. 
After training the network for some time, the accuracy starts to saturate in the later training phase; see label~\circled{2}. 
In this training phase, the notable difference is that smaller batch sizes tend to have better accuracy than bigger batch sizes.
The reason is that, the smaller batch sizes provide more fine-grained information updates than the bigger ones, and considering that the NCARS dataset only has two classes (i.e., car and background), the network tends to train more effectively with small batch sizes while minimizing the overfitting. 
Based on these observations, \textit{we select $B=20$ as the effective value for SNN enhancements, since it quickly leads to stable and high accuracy over the training phase}.
 
%%%%%%%%%%%%%%
\subsection{Impact of the Learning Rate}
\label{Sec_Results_LearnRate}

\begin{figure}[t]
\vspace{-0.1cm}
\centering
\includegraphics[width=0.98\linewidth]{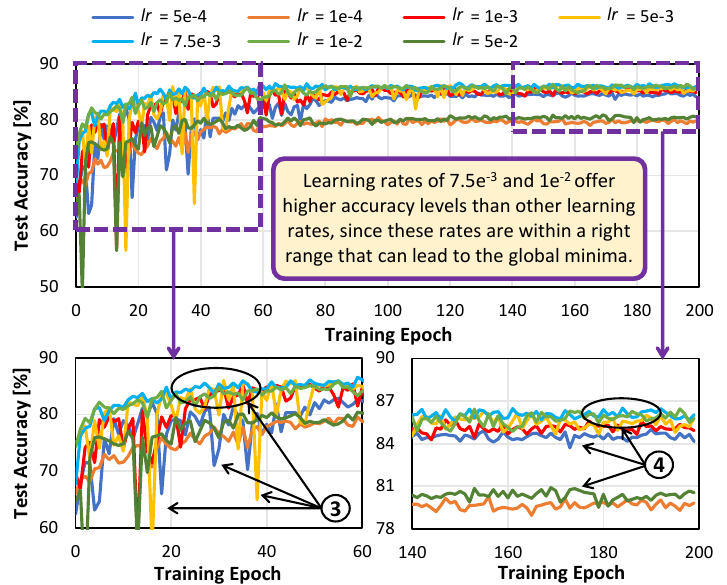}
\vspace{-0.3cm}
\caption{Experimental results for the accuracy across different learning rates, i.e., $lr \in \{5e\text{-}4, 1e\text{-}4, 1e\text{-}3, 5e\text{-}3, 7.5e\text{-}3, 1e\text{-}2, 5e\text{-}2\}$.
Here, we observe that learning rates of $7.5e\text{-}3$ and $1e\text{-}2$ can achieve better accuracy than other learning rates.} 
\label{Fig_Results_LR}
\vspace{-0.6cm}
\end{figure}

The experimental results for the impact of learning rate $lr$ are presented in Fig.~\ref{Fig_Results_LR}.
The results show that, in general, most of the learning rates have fluctuating accuracy during the early training phase (i.e., $\leq$ 60 epoch) since the network still starts learning new information; see label~\circled{3}. 
In this early training phase, we observe that learning rates $7.5e\text{-}3$ and $1e\text{-}2$ can quickly reach accuracy without notable fluctuations, while smaller and bigger learning rates face significant fluctuations.
The reason is that, smaller and bigger learning rates struggle to find local minima, while the learning rates of $7.5e\text{-}3$ and $1e\text{-}2$ are in the right range of values to generalize well.
These trends stay consistent until later training phase, as shown by label~\circled{4}, indicating that the learning rates of $7.5e\text{-}3$ and $1e\text{-}2$ are effective in training the SNN to high accuracy values compared to other learning rates.
To decide which learning rate value to select as the adjustment setting, we compare the accuracy obtained by learning rates $7.5e\text{-}3$ and $1e\text{-}2$. 
Overall, Fig.~\ref{Fig_Results_LR} shows that a learning rate of $7.5e\text{-}3$ achieves comparable accuracy to the learning rate of $1e\text{-}2$ over the training phase, and in the range of 20-40 training epochs, the learning rate of $7.5e\text{-}3$ has slightly better results, meaning that it potentially enables a faster learning process.
Therefore, in this work, \textit{we select $lr = 7.5e\text{-}3$ as the effective value for SNN enhancements}. 
Later, we will also show in Section~\ref{Sec_Results_Discuss} that employing $lr = 1e\text{-}2$ can also achieve good accuracy, demonstrating the effectiveness of our methodology.   

%%%%%%%%%%%%%%
\subsection{Impact of the Threshold Potential}
\label{Sec_Results_Vth}

\begin{figure}[h]
\vspace{-0.3cm}
\centering
\includegraphics[width=0.98\linewidth]{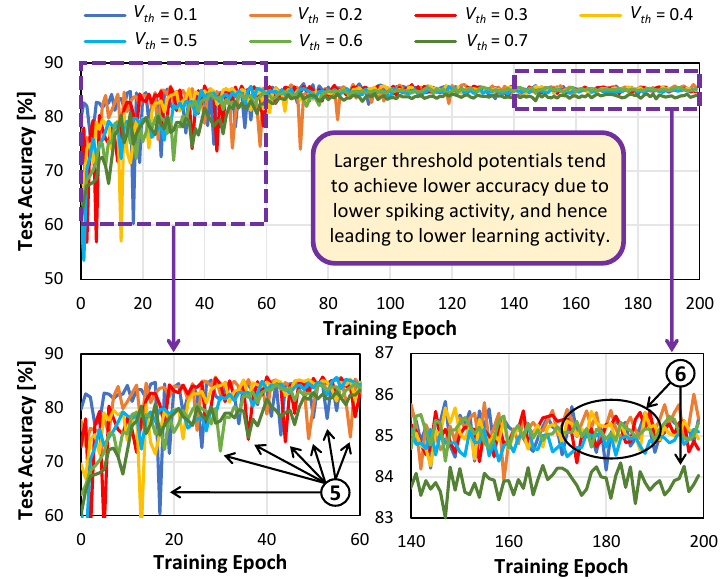}
\vspace{-0.3cm}
\caption{Experimental results for the accuracy across different threshold potentials, i.e., $V_{th} \in \{ 0.1, 0.2, 0.3, 0.4, 0.5, 0.6, 0.7 \}$.
Here, we observe that a threshold potential of $0.7$ achieves lower accuracy than other values.} 
\label{Fig_Results_Vth}
\vspace{-0.4cm}
\end{figure}

The experimental results for the impact of threshold potential $V_{th}$ are presented in Fig.~\ref{Fig_Results_Vth}.
The results show that, most of the given $V_{th}$ values face notable accuracy fluctuations in the early training phase (i.e., $\leq$ 60 epoch) due to the impact of learning new information; see label~\circled{5}.
The accuracy trends become more distinguishable in the later training epochs (i.e., epochs 140-200); see label~\circled{6}.
Here, we observe that most of the threshold potential values (i.e., $0.1, 0.2, 0.3, 0.4, 0.5$, and $0.6$) achieve comparable accuracy, while the threshold potential of $0.7$ has lower accuracy than the others. 
The reason is that a large threshold potential makes the corresponding neuron have low spiking activity, thereby reducing the weight updates for the learning process and limiting the accuracy.   
To decide which threshold potential value to select as the adjustment setting, we consider the one that has minimum significant fluctuations over the training phase. 
Furthermore, the accuracy variation should also be within the range of 1\% accuracy in the later training phase, thereby ensuring a better learning curve over the training phase. 
Based on these criteria, in this work, \textit{we select $V_{th} = 0.5$ as the effective value for SNN enhancements}. 

%%%%%%%%%%%%%%
\subsection{Impact of the Weight Decay}
\label{Sec_Results_Wdecay}

\begin{figure}[t]
\centering
\includegraphics[width=\linewidth]{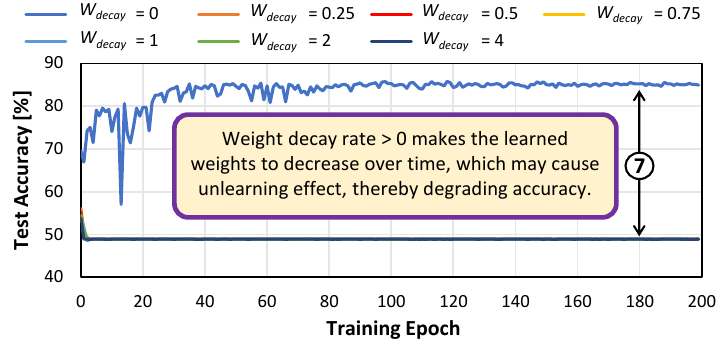}
\vspace{-0.6cm}
\caption{Experimental results for the accuracy across different weight decay rates. i.e., $w_{decay} \in \{ 0, 0.2, 0.5, 0.75, 1, 2, 4 \}$.
Here, we observe that having a weight decay rate $>0$ makes the existing knowledge decay over time and even disappear, thereby degrading the accuracy significantly.} 
\label{Fig_Results_Wdecay}
\vspace{-0.3cm}
\end{figure}

The experimental results for the impact of weight decay $w_{decay}$ are presented in Fig.~\ref{Fig_Results_Wdecay}.
The results show that only a weight decay rate of $0$ offers high accuracy, as in this setting, the knowledge learned by the learning process stays in the network model, thereby providing proper synaptic weights for the inference phase; see label~\circled{7}.
Meanwhile, other weight decay rates (i.e., $>0$) suffer from accuracy degradation, as these settings make the knowledge learned by the learning process decay over time and may disappear eventually, thereby corrupting the synaptic weights for the inference phase; see label~\circled{7}. 
Hence, in this work, \textit{we select $w_{decay} = 0$ as the effective value for SNN enhancements}. 

%%%%%%%%%%%%%%
\subsection{Impact of the SNN Parameter Enhancements }
\label{Sec_Results_Enhance}

\begin{figure}[t]
\centering
\includegraphics[width=0.98\linewidth]{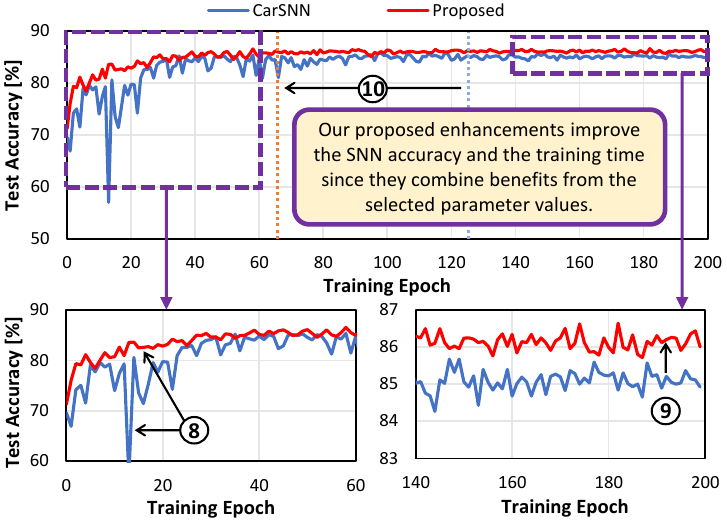}
\vspace{-0.3cm}
\caption{Comparison of accuracy obtained by the state-of-the-art work and our proposed methodology, with the following setting: $B = 20$, $lr = 7.5e\text{-}3$, $V_{th} = 0.5$, and $w_{decay} = 0$. The proposed parameter enhancements lead to higher accuracy than the state-of-the-art work, with a better learning curve over the training phase.} 
\label{Fig_Results_ImprovedCarSNN}
\vspace{-0.6cm}
\end{figure}

Based on previous analysis from Section~\ref{Sec_Results_BatchSize} until Section~\ref{Sec_Results_Wdecay}, we propose a set of parameter values for enhancing SNNs for an event-based automotive dataset: $B = 20$, $lr = 7.5e\text{-}3$, $V_{th} = 0.5$, and $w_{decay} = 0$.
The experimental results for our parameter enhancements are presented in Fig.~\ref{Fig_Results_ImprovedCarSNN}.
The results show that, in general, our parameter enhancements improve the SNN accuracy from the state-of-the-art over the training epoch, demonstrating the effectiveness of our parameter settings.
Our enhanced SNN model has a better accuracy range over the training phase (71.28\% - 86.64\% accuracy) than the state-of-the-art which has a range of 57.06\% - 85.84\% accuracy. 
In the early training phase, the state-of-the-art work faces significant accuracy fluctuations as its parameter values are crafted to achieve stable accuracy after running hundreds of training epochs; see label~\circled{8}. 
Meanwhile, our parameter enhancements can achieve a relatively smooth learning curve (i.e., accuracy) in the transition region, since we craft its parameter values with the ones that iteratively find the local minima over the training phase; see label~\circled{8}.   
These trends stay consistent until the later training phase, since our parameter enhancements achieve notable accuracy improvements as compared to the state-of-the-art work; see label~\circled{9}.
Furthermore, our proposed parameter enhancements can reach a stable region faster than the state-of-the-art work (i.e., by 1.9x speed up). 
Here, the state-of-the-art work achieves 85.24\% accuracy after performing 125 training epochs, while our enhanced SNN model achieves 85.99\% accuracy after performing 66 training epochs; see label~\circled{10}. 
The reason for all these improvements is that we employ parameter enhancements that combine benefits from the selected values, thus maximizing their advantages together for improving SNN accuracy and training time.

%%%%%%%%%%%%%%
\vspace{-0.2cm}
\subsection{Further Discussion}
\label{Sec_Results_Discuss}
\vspace{-0.1cm}

To further demonstrate the effectiveness and generality of our methodology, we perform evaluations considering different parameter settings as the following.
\begin{itemize}
    \item Setting-1: $B = 20$, $lr = 7.5e\text{-}3$, $V_{th} = 0.5$,  $w_{decay} = 0$.
    \item Setting-2: $B = 20$, $lr = 1e\text{-}2$, $V_{th} = 0.5$,  $w_{decay} = 0$.
\end{itemize}

\begin{figure}[t]
\centering
\includegraphics[width=0.97\linewidth]{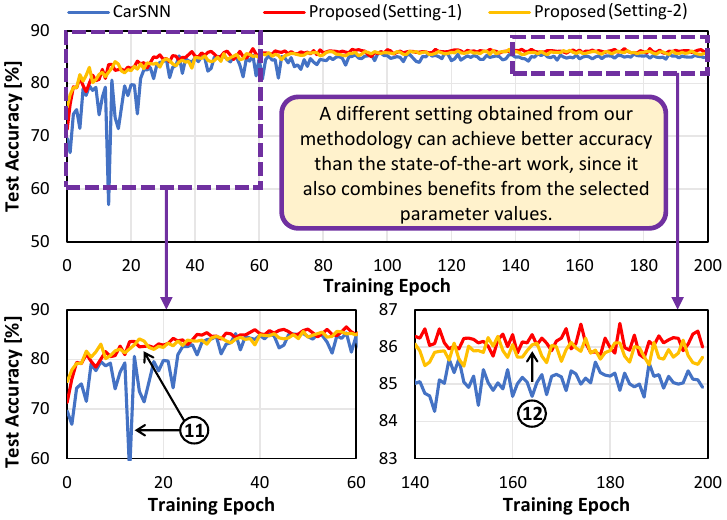}
\vspace{-0.3cm}
\caption{Comparison of accuracy obtained by the state-of-the-art work and our proposed methodology considering different settings, i.e., (1) Setting-1: $B = 20$, $lr = 7.5e\text{-}3$, $V_{th} = 0.5$, $w_{decay} = 0$, and (2) Setting-2: $B = 20$, $lr = 1e\text{-}2$, $V_{th} = 0.5$, $w_{decay} = 0$.} 
\label{Fig_Results_ImprovedDiscuss}
\vspace{-0.6cm}
\end{figure}

Setting-1 is simply the same as our proposed one discussed in the previous subsection (i.e., Section~\ref{Sec_Results_Enhance}), while Setting-2 is similar to Setting-1 but with a different learning rate.
We select a learning rate of $1e\text{-}2$ for Setting-2 because from the observations in Section~\ref{Sec_Results_LearnRate}, this learning rate can achieve comparable accuracy to our selected learning rate value (i.e., $7.5e\text{-}3$). 
The experimental results for this case study are presented in Fig.~\ref{Fig_Results_ImprovedDiscuss}.
The results show that, in general, Setting-2 leads the enhanced SNN model to achieve a better accuracy range over the training phase (i.e., 75.48\% - 86.46\% accuracy) than the state-of-the-art work as indicated by label~\circled{11} and~\circled{12}, demonstrating the effectiveness of its parameter setting. 
Furthermore, Setting-2 can also reach a stable region faster than the state-of-the-art work with 85.17\% accuracy, similar to that of Setting-1.
The reason for all these improvements is that, the SNN parameter values in Setting-2 are also obtained through our methodology, hence combining benefits from the selected parameter values for improving SNN models with event-based autonomous data. 

%%%%%%%%%%%%%%%%%%%%%%%%%%%%%%%%%%%%%%%%%%%%%%%%%%
%%%%%%%%%%%%%%%%%%%%%%%%%%%%%%%%%%%%%%%%%%%%%%%%%%
\vspace{-0.1cm}
\section{Conclusion}
\label{Sec_Conclusion}
\vspace{-0.1cm}

In this work, we propose a novel methodology to systematically understand the impact of SNN parameters on the learning quality, by identifying the practical SNN parameters to explore, and parameter enhancements. We investigate the impact of batch size, learning rate, threshold potential, and weight decay.
The experimental results show that our methodology improves the SNN learning quality, as the enhanced SNN achieves higher accuracy (i.e., 86\%) than the state-of-the-art, and it can speed up the training phase by 1.9x while achieving iso-accuracy (i.e., around 85\% with standard deviation $<$ 0.5\%).
In this manner, our methodology provides effective guidelines and insights for developing highly accurate and efficient SNNs for AD systems. 

%%%%%%%%%%%%%%%%%%%%%%%%%%%%%%%%%%%%%%%%%%%%%%%%%%
%%%%%%%%%%%%%%%%%%%%%%%%%%%%%%%%%%%%%%%%%%%%%%%%%%
\vspace{-0.1cm}
\section*{Acknowledgment}
\label{Sec_Ack}
\vspace{-0.1cm}

This work was partially supported by the NYUAD Center for Interacting Urban Networks (CITIES), funded by Tamkeen under the NYUAD Research Institute Award CG001.

%%%%%%%%%%%%%%%%%%%%%%%%%%%%%%%%%%%%%%%%%%%%%%%%%%
%%%%%%%%%%%%%%%%%%%%%%%%%%%%%%%%%%%%%%%%%%%%%%%%%%
\end{spacing}
\vspace{-0.1cm}

\begin{spacing}{1}
\bibliographystyle{IEEEtran} 
\bibliography{bibliography}

\begin{thebibliography}{10}
\providecommand{\url}[1]{#1}
\csname url@samestyle\endcsname
\providecommand{\newblock}{\relax}
\providecommand{\bibinfo}[2]{#2}
\providecommand{\BIBentrySTDinterwordspacing}{\spaceskip=0pt\relax}
\providecommand{\BIBentryALTinterwordstretchfactor}{4}
\providecommand{\BIBentryALTinterwordspacing}{\spaceskip=\fontdimen2\font plus
\BIBentryALTinterwordstretchfactor\fontdimen3\font minus
  \fontdimen4\font\relax}
\providecommand{\BIBforeignlanguage}[2]{{%
\expandafter\ifx\csname l@#1\endcsname\relax
\typeout{** WARNING: IEEEtran.bst: No hyphenation pattern has been}%
\typeout{** loaded for the language `#1'. Using the pattern for}%
\typeout{** the default language instead.}%
\else
\language=\csname l@#1\endcsname
\fi
#2}}
\providecommand{\BIBdecl}{\relax}
\BIBdecl

\bibitem{Ref_Viale_CarSNN_IJCNN21}
A.~Viale \emph{et~al.}, ``Carsnn: An efficient spiking neural network for
  event-based autonomous cars on the loihi neuromorphic research processor,''
  in \emph{Int. Joint Conf. on Neural Networks (IJCNN)}, 2021, pp. 1--10.

\bibitem{Ref_Cordone_ObjDetSNN_IJCNN22}
L.~Cordone, B.~Miramond, and P.~Thierion, ``Object detection with spiking
  neural networks on automotive event data,'' in \emph{Int. Joint Conf. on
  Neural Networks (IJCNN)}, July 2022, pp. 1--8.

\bibitem{Ref_Shafique_EdgeAI_ICCAD21}
M.~Shafique \emph{et~al.}, ``Towards energy-efficient and secure edge ai: A
  cross-layer framework iccad special session paper,'' in \emph{IEEE/ACM Int.
  Conf. On Computer Aided Design (ICCAD)}, 2021, pp. 1--9.

\bibitem{Ref_Putra_Mantis_ICARA23}
R.~V.~W. Putra and M.~Shafique, ``Mantis: Enabling energy-efficient autonomous
  mobile agents with spiking neural networks,'' in \emph{9th Int. Conf. on
  Automation, Robotics and Applications (ICARA)}, 2023.

\bibitem{Ref_Tavanaei_DLSNN_Neunet18}
A.~Tavanaei \emph{et~al.}, ``Deep learning in spiking neural networks,''
  \emph{Neural Networks}, vol. 111, pp. 47--63, 2019.

\bibitem{Ref_Putra_FSpiNN_TCAD20}
R.~V.~W. {Putra} and M.~{Shafique}, ``Fspinn: An optimization framework for
  memory-efficient and energy-efficient spiking neural networks,'' \emph{IEEE
  Trans. on Computer-Aided Design of Integrated Circuits and Systems (TCAD)},
  vol.~39, no.~11, pp. 3601--3613, 2020.

\bibitem{Ref_Viale_LaneSNN_IROS22}
A.~Viale \emph{et~al.}, ``Lanesnns: Spiking neural networks for lane detection
  on the loihi neuromorphic processor,'' in \emph{2022 IEEE/RSJ Int. Conf. on
  Intelligent Robots and Systems (IROS)}.\hskip 1em plus 0.5em minus
  0.4em\relax IEEE, 2022, pp. 79--86.

\bibitem{Ref_Putra_SpikeDyn_DAC21}
R.~V.~W. Putra and M.~Shafique, ``Spikedyn: A framework for energy-efficient
  spiking neural networks with continual and unsupervised learning capabilities
  in dynamic environments,'' in \emph{58th ACM/IEEE Design Automation
  Conference (DAC)}, 2021, pp. 1057--1062.

\bibitem{Ref_Guo_NeuralCoding_FNINS21}
W.~Guo \emph{et~al.}, ``Neural coding in spiking neural networks: \uppercase{A}
  comparative study for robust neuromorphic systems,'' \emph{Frontiers in
  Neuroscience (FNINS)}, vol.~15, 2021.

\bibitem{Ref_Park_T2FSNN_DAC20}
S.~Park \emph{et~al.}, ``T2fsnn: Deep spiking neural networks with
  time-to-first-spike coding,'' in \emph{57th ACM/IEEE Design Automation
  Conference (DAC)}.\hskip 1em plus 0.5em minus 0.4em\relax IEEE, 2020, pp.
  1--6.

\bibitem{Ref_Chowdhury_LowLatencySNNs_ECCV22}
S.~S. Chowdhury, N.~Rathi, and K.~Roy, ``Towards ultra low latency spiking
  neural networks for vision and sequential tasks using temporal pruning,'' in
  \emph{European Conf. on Computer Vision (ECCV)}.\hskip 1em plus 0.5em minus
  0.4em\relax Springer, 2022, pp. 709--726.

\bibitem{Ref_Putra_TopSpark_IROS23}
R.~V.~W. Putra and M.~Shafique, ``Topspark: A timestep optimization methodology
  for energy-efficient spiking neural networks on autonomous mobile agents,''
  in \emph{2023 IEEE/RSJ Int. Conf. on Intelligent Robots and Systems (IROS)},
  2023, pp. 3561--3567.

\bibitem{Ref_Sironi_HATS_CVPR18}
A.~Sironi \emph{et~al.}, ``Hats: Histograms of averaged time surfaces for
  robust event-based object classification,'' in \emph{IEEE Conf. on Computer
  Vision and Pattern Recognition (CVPR)}, 2018, pp. 1731--1740.

\bibitem{Ruckauer_2019arxiv_NonDifferentiableLossFunctionSNN}
B.~R{\"{u}}ckauer \emph{et~al.}, ``Closing the accuracy gap in an event-based
  visual recognition task,'' \emph{CoRR}, vol. abs/1906.08859, 2019.

\bibitem{Massa_2020IJCNN_EfficientSNNGestures}
R.~Massa \emph{et~al.}, ``An efficient spiking neural network for recognizing
  gestures with a {DVS} camera on the loihi neuromorphic processor,'' in
  \emph{Int. Joint Conf. on Neural Networks (IJCNN)}.\hskip 1em plus 0.5em
  minus 0.4em\relax {IEEE}, 2020, pp. 1--9.

\bibitem{Ref_Neftci_SurrogateSNNs_IEEEMSP19}
E.~O. Neftci, H.~Mostafa, and F.~Zenke, ``Surrogate gradient learning in
  spiking neural networks: Bringing the power of gradient-based optimization to
  spiking neural networks,'' \emph{IEEE Signal Processing Magazine}, vol.~36,
  no.~6, pp. 51--63, 2019.

\bibitem{Ref_Wu_STBP_FNINS18}
Y.~Wu \emph{et~al.}, ``Spatio-temporal backpropagation for training
  high-performance spiking neural networks,'' \emph{Frontiers in Neuroscience
  (FNINS)}, vol.~12, p. 331, 2018.

\bibitem{Ref_Putra_lpSpikeCon_IJCNN22}
R.~V.~W. Putra and M.~Shafique, ``lpspikecon: Enabling low-precision spiking
  neural network processing for efficient unsupervised continual learning on
  autonomous agents,'' in \emph{Int. Joint Conf. on Neural Networks (IJCNN)},
  2022, pp. 1--8.

\bibitem{Ref_Kingma_Adam_ICLR15}
D.~P. Kingma and J.~Ba, ``Adam: A method for stochastic optimization,'' in
  \emph{Int. Conf. on Learning Representations (ICLR)}, 2015.

\bibitem{Ref_Koppula_EDEN_Micro19}
S.~Koppula \emph{et~al.}, ``Eden: Enabling energy-efficient, high-performance
  deep neural network inference using approximate dram,'' in \emph{52nd Annual
  IEEE/ACM Int. Symp. on Microarchitecture (MICRO)}, 2019.

\bibitem{Ref_Putra_SparkXD_DAC21}
R.~V.~W. Putra, M.~A. Hanif, and M.~Shafique, ``Sparkxd: A framework for
  resilient and energy-efficient spiking neural network inference using
  approximate dram,'' in \emph{58th ACM/IEEE Design Automation Conference
  (DAC)}, 2021, pp. 379--384.

\end{thebibliography}
\end{spacing}

\end{document}